\documentclass[runningheads]{llncs}
\usepackage{graphicx}
\usepackage{amsmath}

\begin{document}
\title{Evolutionary algorithms for constructing an ensemble of decision trees}

\author{Evgeny Dolotov\inst{1, 2} \and Nikolai Zolotykh\inst{1, 2}}

%
%\authorrunning{E. Dolotov \and N. Zolotykh}
\institute{Yandex, Moscow, Russia \and National Research University Higher School of Economics, Moscow, Russia\\
\email{\{dolotov-e, nikzolotykh\}@yandex-team.ru}}
\maketitle              % typeset the header of the contribution
\begin{abstract}
Most decision tree induction algorithms are based on a greedy top-down recursive partitioning strategy for tree growth. In this paper, we propose several methods for induction of decision trees and their ensembles based on evolutionary algorithms. The main difference of our approach is using real-valued vector representation of decision tree that allows to use a large number of different optimization algorithms, as well as optimize the whole tree or ensemble for avoiding local optima. Differential evolution and evolution strategies were chosen as optimization algorithms, as they have good results in reinforcement learning problems. We test the predictive performance of this methods using several public UCI data sets, and the proposed methods show better quality than classical methods.

\end{abstract}

\section{Introduction}
Decision trees are a popular method of machine learning for solving classification and regression problems. Because of their popularity many algorithms exists to build decision trees \cite{ref_article1,ref_article2}. However, the task of constructing optimal or near-optimal decision tree is very complex. Most decision tree induction algorithms are based on a greedy top-down recursive partitioning strategy for tree growth. They use different variants of impurity measures, such as information gain \cite{ref_article2}, gain ratio \cite{ref_article4}, gini-index \cite{ref_article5} and distance-based measures \cite{ref_article6}, to select an input attribute to be associated with an internal node. One major drawback of the greedy search is that it usually leads to sub-optimal solutions. The underlying reason is that local decisions at each nodes are in fact interdependent and cannot be found in this way.

A popular approach that can partially solve these problems is the induction of decision trees through evolutionary algorithms (EAs) \cite{ref_article16}. In this approach, each “individual” in evolutionary algorithms represents a solution to the classification problem. Each solution is evaluated by a fitness function, which measures the quality of it. At each new generation, the best solutions have a higher probability of being selected for reproduction. The selected solutions undergo operations inspired by genetics, such as crossover and mutation, producing new solutions which will replace the parents, creating a new population of solutions. This process is repeated until a stopping criterion is satisfied. Instead of a local search, EAs perform a robust global search in the space of candidate solutions. As a result, EAs tend to cope better with attribute interactions than greedy methods and avoid local optima.

In this paper we propose an approach that encodes a decision tree as a real-valued homogeneous vector, since we also encode indices of features by real numbers and decode them using the operation of finding the minimum. This approach allows to use a large number of different optimization algorithms, such as differential evolution \cite{ref_article7} and evolution strategies \cite{ref_article8}.

\section{Related work}
The number of proposed evolutionary algorithms for decision tree induction has grown in the past few years, mainly because they report good predictive accuracy whilst keeping the comprehensibility of decision trees. There are two the most common approaches to encoding decision trees for evolutionary algorithms: tree-based encoding and fixed-length vector encoding. They all use different methods to encode indices of features, threshold values, leaves, and operators in nodes. The main differences in tree-based approaches are the presence of pointers to nodes and the ability to encode trees of various sizes. Axis-parallel decision trees are the most common type found in the literature, mainly because this type of tree is usually much easier to interpret than an oblique tree. A node in axis-parallel decision tree can be described by two parameters: index of tested feature and threshold value. A popular approach \cite{ref_article10} to encoding such trees is to encode each node with one integer and one real number, but in this case, we get heterogeneous and more complex representation of decision tree, than in approach proposed in this article, which makes the process of finding the optimal solution more complex. Authors of article \cite{ref_article11} describe a very similar approach to encoding oblique decision trees with real-valued vectors and optimizing them with differential evolution algorithms, but in this article we propose a more compact representation specifically for axis-parallel decision trees. A more detailed overview of an evolutionary methods for constructing decision trees can be found here \cite{ref_article9}.

\section{Proposed approach}
In this paper we propose a new approach to construct axis-parallel decision tree for classification problems using evolutionary algorithms.
\subsection{Real-valued vector representation}
In axis-parallel trees each node splits dataset according to the following rule:
\begin{equation}
\centering
f(x) =
\begin{cases}
    1, & \text{if } a_i \leq t\\
    0, & \text{otherwise}
\end{cases}
\end{equation}
Thus, each node of the tree is described by two parameters: the index of a feature and a threshold value.

Suppose we have a fixed-length real-valued vector with values in the segment $[0, 1]$. This vector consists of two parts of equal length -- the first part encodes feature indices, and the second part encodes threshold values. Also suppose that all features of objects belong to the segment $[0, 1]$. If this is not the case, then we normalize features using the maximum and minimum values from the training dataset. To restore the index of a feature from the vector, we should find the position of the minimum value in the first part of this vector and find its remainder of integer division by the number of features. The value in the second part of the vector in this position is used as threshold value in the corresponding node. After that, the next minimal value in the vector, the corresponding index of the feature in the node and a threshold value should be found. This operation is repeated until the entire vector is used. Using the indices of features and their threshold values for all nodes, the decision tree without leaves can be built by sequentially adding the nodes. After that, the leaves are added to the decision tree by using training dataset and the majority rule. Thus, we can construct a decision tree from a real-valued vector and evaluate its characteristics.

\subsection{Differential evolution}
The differential evolution (DE)\cite{ref_article7} is an effective evolutionary algorithm designed to solve optimization problems with real-valued parameters. A population in DE consists of $N$ individuals:
\begin{equation}
\centering
P = \{x_1, x_2, ..., x_N\}
\end{equation}
The $j$-th value of the invidual $x_i$ in the initial population is calculated as follows:
\begin{equation}
\centering
x_j^i = x_j^{\min} + r(x_j^{\max} - x_j^{\min}),
\end{equation}
where $r \in [0, 1]$ is a uniformly distributed random number.

The evolutionary process implements an iterative scheme to evolve the initial population. At each iteration of this process, known as the generation, a new population of individuals is generated from the previous one. Each individual is used to build a new vector by applying the mutation and crossover operators:
\begin{itemize}
    \item Mutation. Three randomly chosen individuals are linearly combined as follows:
    \begin{equation}
    \centering
    v^i = x^{j_1} + \alpha(x^{j_2} - x^{j_3}),
    \end{equation}
    where $\alpha$ is a user-specified constant.
    \item Crossover. The mutated vector is recombined with the target vector to build the trial vector:
    \begin{equation}
    \centering
    u_j^i =
    \begin{cases}
        v_j^i, & \text{if } r \leq CR \text{ or } j=l\\
        x_j^i, & \text{otherwise}
    \end{cases}
    \end{equation}
    where $r \in [0,1]$ is uniformly distributed random number and $CR$ is the crossover rate.
    \item Selection: A one-to-one tournament is applied to determine which individual is selected as a member of the new population.
\end{itemize}

In the final step, when a stop condition is fulfilled, DE returns the best individual in the current population.
\subsection{Evolution strategies}
Unlike the method of differential evolution, the population in the method of evolution strategies \cite{ref_article8} consists of only one individual:
\begin{equation}
\centering
P \sim x
\end{equation}
The initial individual is calculated as follows:
\begin{equation}
\centering
x_j = x_j^{\min} + r(x_j^{\max} - x_j^{\min}),
\end{equation}
where $r \in [0, 1]$ is a uniformly distributed random number.
We sample several offsets which are represented as a normal distributed random vector $e_1, e_2, ..., e_n \sim \mathcal{N}(0, I)$. Then we shift the individual in the direction of the weighted sum of the offsets, which approximate the gradient:
\begin{equation}
\centering
x \leftarrow x + \alpha\frac{1}{n\sigma}\sum_{i=1}^{n}f(x+\sigma e_i)e_i,
\end{equation}
where $\alpha$ and $\sigma$ are user-specified constants.
\subsection{Construction of ensembles}
Two of the most popular approaches for constructing the ensembles of decision trees is bagging and boosting. Example of method that use bagging approach is random forest and example of method that use boosting approach is AdaBoost. In this part of the paper we propose to replace classical algorithms to induction of decision trees in these methods by evolutionary algorithms described earlier. Thus, we obtain two new methods: evolutionary random forest (EvoRF) as the analogue of random forest and EvoBoost as the analogue of AdaBoost. In addition to this, we consider the method (EvoEnsemble) in which each individual in the population is the whole ensemble, and representation of the ensemble is a large real-valued vector obtained by concatenation with a vector for each tree from the ensemble. Thus, in this method -- evolutionary ensemble, we optimize the whole ensemble at once, which theoretically should lead to a better result.

\section{Experiment}
For experiments we use several popular datasets from UCI repository. Experiments are divided into two parts.

First, we evaluate classification accuracy of the methods based on evolutionary algorithms and compare their results with the classical methods for solve classification problems. Experiments show that using of the proposed methods does not allow to exceed the results of classical algorithms for constructing decision trees on some datasets, but on the vast majority of datasets using evolution strategies allows to achieve the significant improvement in the accuracy of prediction by several percent (Table \ref{tab1}). Therefore, we decide to use this algorithm to build ensembles in subsequent experiments.

\begin{table}
\caption{Comparison of popular classification algorithms such as CART \cite{ref_article12} and multilayer perceptron (MLP) \cite{ref_article13} with the proposed approaches: differential evolution (DE) and evolution strategies (ES).}\label{tab1}
\centering
\begin{tabular}{|c|c|c|c|c|c|c|c|c|c|}
\hline
\textbf{Dataset} & \textbf{CART} & \textbf{MLP} & \textbf{DE} & \textbf{ES} & \textbf{Dataset} & \textbf{CART} & \textbf{MLP} & \textbf{DE} & \textbf{ES}\\
\hline
car & 96.74 & \textbf{98.32} & 90.59 & 91.18 & molecular-p & 75.85 & \textbf{86.54} & 85.57 & 86.01\\
\hline
tic-tac-toe & \textbf{93.65} & 93.38 & 87.96 & 86.39 & diabets & 74.49 & 73.89 & 75.03 & \textbf{75.07}\\
\hline
glass & 71.42 & 71.36 & 73.02 & \textbf{73.45} & balance-scale & 78.07 & 79.68 & \textbf{80.05} & 80.04\\
\hline
iris & 94.45 & 97.13 & 96.97 & \textbf{97.24} & ionosphere & 88.23 & 89.78 & \textbf{91.32} & 91.17\\
\hline
australian & 85.67 & 86.12 & \textbf{86.42} & 86.05 & cmc & 54.83 & \textbf{56.05} & 55.89 & 56.01\\
\hline
wine & 92.43 & 93.75 & 94.58 & \textbf{94.65} & vehicle & 69.75 & \textbf{72.31} & 71.96 & 72.18\\
\hline
liver-disoder & 67.73 & 66.96 & \textbf{68.36} & 68.25 & lympth & 77.97 & 78.13 & \textbf{78.42} & 78.36\\
\hline
haberman & 73.25 & 74.89 & 75.43 & \textbf{75.76} & dermatology & 94.32 & 93.56 & 95.67 & \textbf{95.75}\\
\hline
heart-statlog & 78.75 & 75.43 & 79.34 & \textbf{80.20} & sonar & 75.33 & 77.35 & 76.49 & \textbf{79.43}\\
\hline
page-blocks & 96.98 & 95.76 & \textbf{97.35} & 97.03 & credit-g & 72.25 & \textbf{75.43} & 74.32 & 73.85\\
\hline
\end{tabular}
\end{table}

Second, we evaluate classification accuracy of several approaches to constructing ensembles of decision trees. For these experiments we use datasets which have only two different labels, in other words, we are solving the problem of binary classification. Various hyperparameter of the random forest algorithm and AdaBoost such as, the depth of trees and maximum number of trees was selected using the method of grid search, and then these same parameters were used for their evolutionary analogues. As well as in the case of using evolutionary algorithms for constructing a single decision tree, experiments show that using of the proposed methods does not allow to exceed the results of classical algorithms for constructing ensemble of decision trees on some datasets, but the method that represent whole ensemble as one real-valued vector are showing best accuracy on most datasets (Table \ref{tab2}).

\begin{table}
\caption{Comparison of random forest (RF) \cite{ref_article14} and AdaBoost \cite{ref_article15} with the proposed approaches for constructing the ensembles of decision trees: evolutionary version of random forest (EvoRF), AdaBoost, EvoBoost and evolution ensemble (EvoEnsemble)}\label{tab2}
\centering
\begin{tabular}{|c|c|c|c|c|c|}
\hline
\textbf{Dataset} & \textbf{RF} & \textbf{EvoRF} & \textbf{EvoEnsemble} & \textbf{AdaBoost} & \textbf{EvoBoost}\\
\hline
tic-tac-toe & 97.48 & 97.76 & \textbf{97.84} & 96.31 & 96.91\\
\hline
australian & 92.03 & 91.59 & \textbf{92.73} & 91.36 & 90.93\\
\hline
liver-disoder & \textbf{77.32} & 75.27 & 76.73 & 76.31 & 76.45\\
\hline
molecular-p & 89.24 & 90.64 & \textbf{91.03} & 90.21 & 90.40\\
\hline
diabets & 82.31 & 83.74 & 83.67 & 82.23 & \textbf{85.07}\\
\hline
ionosphere & 92.35 & 92.89 & \textbf{93.11} & 91.76 & 92.17\\
\hline
haberman & 79.45 & 80.12 & 80.79 & 79.21 & \textbf{80.69}\\
\hline
heart-statlog & 83.43 & 84.24 & 83.78 & 83.09 & \textbf{83.85}\\
\hline
sonar & 85.14 & 86.38 & \textbf{86.19} & 85.02 & 86.03\\
\hline
credit-g & 77.31 & 77.24 & 79.15 & 79.07 & \textbf{79.63}\\
\hline
\end{tabular}
\end{table}

\section{Conclusion and Future Work}
In this paper, we have proposed several methods that use different evolutionary algorithms to construct decision trees and their ensembles. The main contribution of this paper is method to construct real-valued vector representation of decision tree that allows to use different evolutionary algorithms for constructing decision trees and their ensembles. The proposed algorithms show better quality than classical methods such as CART, random forest and AdaBoost on popular datasets from UCI repository, but in order to achieve such high results, it takes more time than using classical algorithms. This is due to the fact that the methods using evolutionary algorithms during training several times build trees and evaluate their quality, while classical algorithms do it only once.

A detailed analysis of the computational performance of the proposed methods, parallel computations in evolutionary algorithms, initialization of initial population by results of the classical decision tree inductions algorithms and evolutionary analogue of gradient boosting are possible areas for further research.


\begin{thebibliography}{99}
\bibitem{ref_article1} R. O. Duda, P. E. Hart, and D. G. Stork. Pattern classification, 2nd ed., Wiley interscience, 2001.

\bibitem{ref_article2} J. R. Quinlan. Induction of decision trees, Machine Learning, vol. 1, no. 1, pp. 81-106, 1986.

\bibitem{ref_article4} J.R. Quinlan. C4.5: programs for machine learning. Morgan Kaufmann Publishers Inc., 1993.

\bibitem{ref_article5} L. Breiman, J. H. Friedman, R. A. Olshen, and C. J. Stone, Classification and Regression Trees, 1984.

\bibitem{ref_article6} R. L. De M\'antaras, A distance-based attribute selection measure for decision tree induction, Machine Learning, vol. 6, no. 1, pp. 81–92, 1991.

\bibitem{ref_article7} M. Tasgetiren, Y. Liang, M. Sevkli, and G. Gencyilmaz, Differential evolution algorithm for permutation flowshop sequencing problem with makespan criterion. In Proceedings of the 4th International Symposium on Intelligent Manufacturing Systems(IMS2004), pp. 442-452, 2004.

\bibitem{ref_article8} I. Rechenberg and M. Eigen. Evolutionsstrategie: Optimierung Technischer Systeme nach Prinzipien der Biologischen Evolution. Frommann-Holzboog Stuttgart, 1973.

\bibitem{ref_article10} D. Jankowski and K. Jackowski. K. Evolutionary algorithm for decision tree induction. In IFIP International Conference on Computer Information Systems and Industrial Management, pp. 23-32, 2015.

\bibitem{ref_article11} R. Rivera-Lopez and J. Canul-Reich. Differential Evolution Algorithm in the Construction of Interpretable Classification Models. In Artificial Intelligence-Emerging Trends and Applications, 2018.

\bibitem{ref_article9} M.P. Basgalupp, A. Carvalho, R.C. Barros, A. Freitas. A Survey of Evolutionary Algorithms for Decision-Tree Induction. IEEE Transaction on Systems, Man and Cybernetics, Part C (Applications and Reviews), 2012.

\bibitem{ref_article12} L. Breiman, J.H. Friedman, R.A. Olshen, and C.J. Stone. Classification and Regression Trees, Wadsworth, Belmont, CA. Republished by CRC Press, 1984.

\bibitem{ref_article13} S. Haykin. Neural networks: a comprehensive foundation. Prentice Hall PTR, 1994.

\bibitem{ref_article14} L. Breiman. Random forests. Machine learning, vol. 45, no. 1, pp 5-32, 2001.

\bibitem{ref_article15} Y. Freund and R. E. Schapire. A decision-theoretic generalization of on-line learning and an application to boosting. Journal of computer and system sciences, vol. 55, no.1, pp 119-139, 2001.

\bibitem{ref_article16} T. Back. Evolutionary algorithms in theory and practice: evolution strategies, evolutionary programming, genetic algorithms. Oxford university press, 1996.

\end{thebibliography}
\end{document}